\documentclass[conference]{IEEEtran}
\IEEEoverridecommandlockouts
\usepackage{subcaption}
\usepackage{algorithm}
\usepackage{algpseudocode}
\usepackage{cite}
\usepackage{amsmath,amssymb,amsfonts}

\usepackage{graphicx}
\usepackage{textcomp}
\usepackage{xcolor}
\usepackage{bbm}

\def\BibTeX{{\rm B\kern-.05em{\sc i\kern-.025em b}\kern-.08em
    T\kern-.1667em\lower.7ex\hbox{E}\kern-.125emX}}
\begin{document}

\title{Quantifying Calibration Error in Neural Networks through
Evidence-Based Theory\\
\thanks{We thank the partners in the European CONNECT project for their fruitful discussions on trust management. This work is partially funded by the HORIZON CONNECT project under EU grant agreement no. 101069688}
}

\author{\IEEEauthorblockN{Koffi Ismael Ouattara}
\IEEEauthorblockA{\textit{Huawei Technologies} \\
Munich, Germany \\
koffi.ismael.ouattara@huawei.com}
\and
\IEEEauthorblockN{Ioannis Krontiris}
\IEEEauthorblockA{\textit{Huawei Technologies} \\
Munich, Germany \\
ioannis.krontiris@huawei.com}
\and
\IEEEauthorblockN{Theo Dimitrakos}
\IEEEauthorblockA{\textit{Huawei Technologies} \\
Munich, Germany \\
theo.dimitrakos@huawei.com}
\and
\IEEEauthorblockN{Frank Kargl}
\IEEEauthorblockA{Ulm University, Institute of \\ Distributed Systems\\
frank.kargl@uni-ulm.de}}

\maketitle

\begin{abstract}
Trustworthiness in neural networks is crucial for their deployment in critical applications, where reliability, confidence, and uncertainty play a pivotal role in decision-making. Traditional performance metrics such as accuracy and precision fail to capture these aspects, particularly in cases where models exhibit overconfidence. To address these limitations, this paper introduces a novel framework for quantifying the trustworthiness of neural networks by incorporating subjective logic into the evaluation of Expected Calibration Error (ECE). This method provides a comprehensive measure of trust, distrust, and uncertainty by clustering predicted probabilities and fusing opinions using appropriate fusion operators. We demonstrate the effectiveness of this approach through experiments on the MNIST and CIFAR-10 datasets, where post-calibration results indicate improved trustworthiness. The proposed framework offers a more interpretable and nuanced assessment of AI models, with potential applications in sensitive domains such as healthcare and autonomous systems.
\end{abstract}


\section{Introduction}
Artificial Intelligence (AI) systems, particularly neural networks, are increasingly employed in critical applications such as healthcare, finance, and autonomous systems. These systems play an integral role in decision-making, where the trustworthiness of their predictions becomes paramount. Trustworthiness in AI encompasses attributes such as reliability, robustness, fairness, and transparency, yet these qualities are often difficult to evaluate, particularly in neural networks, which are typically viewed as "black-box" models. This opacity raises significant concerns about their trustworthiness, especially in sensitive domains where incorrect decisions can lead to severe consequences.

Traditional performance metrics like accuracy, precision, and recall measure only the correctness of the model’s predictions but fail to capture the confidence and uncertainty associated with those predictions. Confidence calibration, which aligns predicted probabilities with actual outcomes, has emerged as an important tool to address these shortcomings. Well-calibrated models provide predictions where the predicted probability corresponds to the actual likelihood of the event, ensuring that a 70\% confidence means the event occurs approximately 70\% of the time. However, despite its utility, calibration alone does not fully address the issue of trustworthiness, as it does not account for subjective uncertainty or provide an interpretable way to assess trust across a range of predictions.

To address this, we propose the use of subjective logic for trustworthiness quantification in neural networks. Subjective logic extends probabilistic logic by incorporating degrees of belief, disbelief, and uncertainty, making it well-suited for analyzing trust in uncertain environments. By combining subjective logic with existing calibration techniques, we aim to provide a more nuanced and interpretable framework for evaluating the trustworthiness of neural network models.

\emph{Problem.}
Evaluating the trustworthiness of an AI system, particularly neural networks, poses significant challenges. Traditional metrics like accuracy, precision, and recall do not capture the uncertainties and subjective aspects of trustworthiness. Confidence calibration methods, such as those discussed in~\cite{guo2017calibrationmodernneuralnetworks}, introduce the Expected Calibration Error (ECE) to align predicted probabilities with actual outcomes, aiming to make AI systems more trustworthy. However, ECE, while useful, is not easily interpretable and does not provide a comprehensive view of an AI system's trustworthiness.

\emph{Contribution.}
To address these challenges, we propose a novel scheme for trustworthiness quantification. Building on the concept of Expected Calibration Error (ECE), our scheme evaluates the trustworthiness of classification AI models by clustering predicted probabilities and computing trustworthiness opinions for each cluster. This approach provides a detailed and intuitive trustworthiness assessment, which can be fused into a single trust opinion using appropriate fusion operators. By integrating subjective logic into this process, our method offers a more comprehensive and interpretable framework for assessing the trustworthiness of neural networks, crucial for their deployment in critical applications.

\emph{Structure.}
This paper is organized as follows. Section 2 reviews related work, discussing previous research on AI trustworthiness and classifying them. Section 3 provides a background, detailing subjective logic, its operators, and neural network calibration. Section 4 outlines the problem statement, highlighting the challenges in quantifying AI trustworthiness, especially in the context of neural networks. Section 5 describes our contribution, elaborating on the proposed method, the process of clustering predicted probabilities, and computing trustworthiness opinions. Finally, Section 6 presents the evaluation, including empirical analysis and results, demonstrating the effectiveness of our approach in real-world scenarios.

\emph{Summary.}
By incorporating subjective logic into the trustworthiness evaluation of neural networks, our approach not only addresses the limitations of traditional metrics but also provides a more comprehensive and interpretable framework for assessing AI trustworthiness. This is crucial for enhancing the reliability and ethical deployment of AI systems in critical applications.

\section{Related Work}
The quantification of AI trustworthiness is an increasingly critical area of research, driven by the imperative to ensure that AI systems are reliable, fair, and interpretable. This section reviews key works on evaluating trust in neural networks (NNs).

\textbf{Interpretable Trust Quantification Metrics for NNs}~\cite{wong2021reallytrustyousimple}: proposes a novel set of interpretable metrics aimed at quantifying the trustworthiness of NNs based on their performance in a series of questions. Their research introduces metrics such as Question-Answer Trust, Trust Density, Trust Spectrum, and NetTrustScore. These metrics focus on the \emph{accuracy and confidence} of AI outputs, emphasizing the importance of output confidence in the broader context of AI reliability without delving into other aspects like privacy or security. Their approach is \emph{static}, as it establishes trust in the AI system through a one-time evaluation. It is a \emph{black box} method, focusing on the AI system's outputs without requiring insight into its internal processes. 

\textbf{Comprehensive Framework for Trustworthiness Evaluation}~\cite{zora255635}: Contrasting with the singular focus on output confidence, \cite{zora255635} offers a broader evaluation encompassing \emph{fairness, explainability, robustness, and accountability}. This framework provides a structured methodology for a holistic assessment of trust in AI systems, underlining the importance of these four pillars in establishing a comprehensive understanding of AI trustworthiness. Their approach is \emph{static} and does not explicitly align with the \emph{white or black box} paradigms but offers a structured evaluation that could incorporate elements of both. 

\textbf{Survey on Uncertainty in NNs}~\cite{gawlikowski2022surveyuncertaintydeepneural}: focuses on the quantification of uncertainty inherent in NN predictions. By categorizing uncertainty into model, data, and distributional types, their work highlights the importance of distinguishing these uncertainties to enhance the reliability and trustworthiness of AI systems in high-stakes applications. The survey underscores the challenge of establishing trust based solely on output \emph{accuracy/confidence}. 
The paper delves into "Single Deterministic Methods," exploring a dynamic and \emph{white box} approach for assessing uncertainty in neural network (NN) predictions. This method involves utilizing single deterministic network methods to quantify uncertainty through a single forward pass within the network, enabling \emph{dynamic} evaluation of uncertainty in each prediction. Emphasizing confidence in output accuracy over other aspects of AI trustworthiness like privacy or security, it adopts internal uncertainty quantification approaches to emulate Bayesian modeling for quantifying model uncertainty. By directly assessing uncertainty from the network's output, this methodology enhances interpretability and reliability of NN outputs, contributing significantly to discussions on AI trustworthiness, particularly in contexts where accurate and confident predictions are crucial.

\textbf{Evidential Deep Learning for Uncertainty Quantification}~\cite{sensoy2018evidentialdeeplearningquantify}: introduces a novel framework that leverages subjective logic for direct modeling of uncertainty in NN predictions. By treating predictions as subjective opinions and using Dirichlet distribution for uncertainty quantification, their method provides a \emph{dynamic} and interpretable approach that improves robustness against adversarial perturbations and out-of-distribution queries. Like previous work, this one is also a \emph{white-box} approach and focusing only on uncertainty (\emph{accuracy/confidence}). The main difference compared to previous work is that this work is based on \emph{Subjective Logic} framework.

\textbf{DeepTrust Framework for Trust Quantification in NNs}~\cite{10.3389/frai.2020.00054}: proposes DeepTrust which uses \emph{Subjective Logic} to quantify trust in neural networks. DeepTrust is notable for its \emph{dynamic} evaluation of trust, both in individual predictions and the neural network as a whole. By incorporating the trustworthiness of the dataset and the algorithmic processes, DeepTrust offers a nuanced assessment of AI trustworthiness. In a nutshell the solution proposed is a \emph{white box} approach and is dynamic since it allows to evaluate trust in each prediction of the NN but still, it also allows to evaluate only trust in the NN. 

Unlike some of the methods reviewed, our approach is static, as it evaluates the trustworthiness of the neural network as a whole rather than dynamically assessing the trustworthiness of individual predictions or uncertainties. By adopting a black-box approach (meaning that the AI to assess is seen as a black box), our method does not require insight into the internal workings of the neural network, making it broadly applicable to various types of models. Furthermore, our primary focus is on improving the accuracy of predicted probabilities through calibration, specifically addressing the alignment between model confidence and actual outcomes. While subjective logic is used to provide an intuitive and interpretable trustworthiness assessment, the emphasis remains on ensuring that the overall confidence of the network is reliable for deployment in critical applications.


\section{Background}
\subsection{Subjective Logic}
\subsubsection{Subjective Opinion}

Subjective logic~\cite{josang2016subjective} is an extension of probabilistic logic that incorporates a second level of uncertainty. This additional layer allows for expressions such as "I don't know," enabling a random variable to capture uncertainty more explicitly. Moreover, subjective logic allows an agent to express an opinion about a given proposition. The notation for a subjective opinion is $\omega_X^A$. The subscript $X \in \mathbb{X}$ indicates the target variable or proposition to which the opinion applies, and the superscript $A$ indicates the agent (i.e., the belief owner) who holds the opinion. In Subjective Logic, depending on $\mathbb{X}$ we can distinguish three types of opinions: binomial, multinomial and hypernomial. For our work, only the binomial type is relevant. 

In binomial opinion, the cardinal of the domain \(\mathbb{X}\) is 2 \(\left(\{x, \bar{x}\}\right)\) and the opinion has the form $\omega_X^A = (b_x, d_x, u_x, a_x)$, where
$b_x$ is the belief mass in support of $x$ being TRUE (i.e. $X = x$ ), $d_x$ is the disbelief mass in support of $X$ being FALSE (i.e. $X = \bar{x}$ ), $u_x$ is the uncertainty mass representing the vacuity of evidence, and $a_x$ is the base rate, i.e. prior probability of $x$ without any evidence.
These parameters follow the constraint \(b_x + d_x + u_x = 1\), ensuring the total belief, disbelief, and uncertainty add up to unity. 

The \textit{projected probability} of the opinion is then defined as:
\(
P(x) = b_x + u_x \cdot a_x
\)

This form allows for a flexible representation of both the evidence and the lack of evidence for a particular hypothesis. 

Equation~\ref{eq:1} establishes a bijective mapping between a binomial opinion and the Beta probability density function (Probability of probability). This Beta Probability Density Function, denoted as \( \text{Beta}^e(p_x, r_x, s_x, a_x) \), is described by two shape parameters \(r_x\) and \(s_x\), and a base rate \(a_x\). The shape parameters \(r_x\) and \(s_x\) control the shape of the Beta distribution, while \(a_x\) represents the prior belief about the binomial event in the absence of data. \(r_x\) and \(s_x\) can be interpreted as, respectively, the number of positive and negative evidence for \(X\) taking the value \(x\).


The bijective mapping is as follows~\cite{josang2016subjective}:
\begin{equation}
\label{eq:1}
\begin{aligned}
    b_x &= \frac{r_x}{W + r_x + s_x}, \\
    d_x &= \frac{s_x}{W + r_x + s_x}, \\
    u_x &= \frac{W}{W + r_x + s_x},
\end{aligned}
\quad
\Longleftrightarrow
\quad
\begin{aligned}
    r_x &= \frac{b_x W}{u_x}, \\
    s_x &= \frac{d_x W}{u_x}, \\
    1 &= b_x + d_x + u_x.
\end{aligned}
\text{for } \, u_x\neq0 
\end{equation}

where \(W\) is a weight parameter that controls the amount of vacuous evidence. In the case of \textit{vacuous opinions} where no prior evidence exists, the non-informative default weight \(W = 2\) is commonly used, which corresponds to a uniform Beta distribution, \( \text{Beta}(p_x, 1, 1) \). More binomial opinion quantification method can be found in another of our submitted paper~\cite{dataai}.

\subsubsection{Analyze Trust with Subjective Logic}

Subjective logic can be used to analyze trust~\cite{10.5555/1151699.1151710, 10706345}. \cite{josang2016subjective} defines trust as a subjective binomial opinion $(t, d, u)$\footnote{The base rate is usually always set to \(0.5\).}, where \(t\) is the belief mass (or trust mass), \(d\) is the disbelief mass (or distrust mass), and \(u\) is the uncertainty (or untrust mass). It also introduced fusion operators that can be used to fuse information derived using different ways. These fusion operators are essential for merging opinions in various scenarios, such as trust analysis and decision-making.

Several fusion operators are used depending on the nature of the information and the relationship between sources. Belief Constraint Fusion (BCF) applies when no compromise is possible between opinions, meaning no conclusion is drawn if there is total disagreement. Cumulative Belief Fusion (CBF), in its aleatory and epistemic forms, assumes that adding more evidence reduces uncertainty, especially in statistical processes (A-CBF) or subjective knowledge (E-CBF). Averaging Belief Fusion (ABF) is used when opinions are dependent but equally valid, averaging them without assuming more evidence increases certainty. Weighted Belief Fusion (WBF) gives more confident opinions greater weight, ideal for expert input where confidence varies. Finally, Consensus \& Compromise Fusion (CCF) preserves shared beliefs while turning conflicting opinions into vague beliefs, reflecting uncertainty and fostering consensus.

Choosing the appropriate fusion operator depends on the specific situation. For example, BCF is useful when strict agreement is required, while CCF is suited for cases where compromise is possible. By understanding the nature of the opinions and their relationships, analysts can select the most effective fusion operator to ensure accurate and meaningful results.
\subsection{Neural Network Calibration}

Neural networks have become the backbone of many modern AI applications due to their ability to learn and generalize from large datasets. However, an important aspect of their performance that has received increasing attention is calibration. Calibration refers to the degree to which a model's predicted probabilities align with the true likelihood of outcomes. A well-calibrated model provides predicted probabilities that match the actual frequencies of events. For example, among all instances where a model predicts a 70\% probability of an event, that event should occur approximately 70\% of the time.

Despite their high accuracy, many state-of-the-art neural networks, such as deep convolutional networks used in image classification or recurrent networks in natural language processing, often suffer from poor calibration~\cite{wang2024calibration}. This misalignment can lead to overconfident or underconfident predictions, which is problematic in applications where uncertainty estimation is crucial, such as medical diagnosis or autonomous driving.

\subsubsection{Calibration Error}
Expected Calibration Error (ECE) is one of the most common metrics used to quantify calibration quality. ECE compares the predicted confidence of a model with the actual outcomes by grouping predictions into bins based on confidence levels. Each bin represents a range of predicted probabilities (e.g., 0.0 to 0.1, 0.1 to 0.2, etc.). The difference between the model’s average accuracy in each bin and its average predicted confidence gives a measure of the miscalibration. Formally, ECE is calculated as:
\[
ECE = \sum_{m=1}^M \frac{|B_m|}{n} \left| \text{acc}(B_m) - \text{conf}(B_m) \right|
\]

where \(M\) is the number of bins, \(B_m\) represents the set of predictions in bin \(m\), and \(n\) is the total number of samples. Perfect calibration occurs when the predicted confidence exactly matches the actual accuracy in every bin, resulting in an ECE of 0. A higher ECE indicates miscalibration, which could manifest as overconfidence (the predicted probability is higher than the actual outcome) or underconfidence (the predicted probability is lower than the actual outcome).

While useful, ECE does not offer a full view and interpretable quantity of a model’s trustworthiness.

\subsubsection{Calibration Techniques}

Several methods have been developed to improve neural network calibration. One of the most effective and widely-used techniques is Temperature Scaling, a simple post-processing method introduced by \cite{guo2017calibrationmodernneuralnetworks}. In this method, the logits\footnote{The logits are the raw outputs of the neural network before applying the softmax function} are divided by a scalar temperature parameter \(T\), which is optimized using a validation set. The modified logits are then fed into the softmax function to generate calibrated probabilities. The temperature parameter 
\(T\) is tuned to reduce the difference between predicted probabilities and actual outcomes, as measured by metrics like the Negative Log-Likelihood (NLL).

Temperature scaling does not affect the model’s classification accuracy but reduces its confidence in predictions, resulting in better-calibrated probabilities. This technique is particularly advantageous because it is easy to implement, computationally inexpensive, and does not require retraining the model. By adjusting the temperature, it makes overconfident models more realistic in their probability estimates, thus improving their trustworthiness.

\section{Problem statement}
The increasing deployment of neural networks in critical applications such as healthcare, finance, and autonomous systems necessitates the need for reliable and ethical decision-making. Despite their high performance in terms of accuracy, these neural networks are often perceived as black-box systems due to their complex and opaque nature. This opacity leads to significant challenges in evaluating the trustworthiness of these models, which is crucial for their adoption in sensitive domains.

Traditional metrics like accuracy, precision, and recall focus solely on the performance of neural networks in terms of their output but fail to capture the uncertainties associated with their predictions. These metrics do not provide insights into how much confidence one can place in the model's predictions, particularly in situations where the model might be overconfident or underconfident.

One approach to improving trustworthiness is through confidence calibration, where methods like the Expected Calibration Error (ECE) are used to evaluate how well predicted probabilities are aligned with actual outcomes. However, while ECE provides a useful measure of calibration, it is not easily interpretable and does not offer a comprehensive view of an AI system’s trustworthiness. ECE focuses on the difference between predicted probabilities and actual outcomes but lacks the capability to represent the degrees of belief, disbelief, and uncertainty that are crucial for a more nuanced trust assessment.


\section{Quantification Framework}
\begin{algorithm}
\caption{Overall Process for Trustworthiness Evaluation}
\begin{algorithmic}[1]
    \State \textbf{Input:} Predicted probabilities from a neural network for each class for a set of input
    \State \textbf{Output:} Trust opinion on the neural network
    
    \State \textbf{Step 1: Clustering Predicted Probabilities}
    \State Cluster the predicted probabilities into \(M\) groups based on their values within the range $[0, 1]$.
    \State For each cluster, assign a representative value (e.g., the middle of the interval representing the cluster).

    \State \textbf{Step 2: Computing Trustworthiness Opinions for each class}
    \For{each class $c$}
        \State \textbf{Step 2.1: Computing Trustworthiness Opinions for each cluster}
        \For {each cluster $i$}
            \State Set $RP_i$ to the representative of the cluster
            \State Calculate the number of classifications $n_i$ falling into this cluster.
            \State Calculate the number of good classification $t_i$ in this cluster.
            \State Set the number of Positive Evidence $r$ to $t_i$
            \State Set the number of negative evidence $s$ to $\lvert t_c - n_c*RP_i\rvert$
            \State Compute trust opinion for the cluster using subjective logic binomial opinion quantification.
        \EndFor
        \State Fuse the trust opinions from all clusters using fusion operators to obtain a single trust opinion for the class.
    \EndFor
    
    \State \textbf{Step 3: Final Trust Opinion on Neural Network}
    \State Fuse the trust opinions from all classes to obtain the final trust opinion for the neural network.
\end{algorithmic}
\label{alg:1}
\end{algorithm}

This paper tackles the challenges of quantifying the trustworthiness of neural networks by introducing a novel framework grounded in subjective logic. Our primary contribution lies in developing a comprehensive trustworthiness quantification method that extends the traditional Expected Calibration Error (ECE). This framework incorporates subjective measures of belief (trust), disbelief (distrust), and uncertainty, offering a more nuanced and interpretable evaluation of model trustworthiness.
The framework operates through the following steps summarized in Algorithm~\ref{alg:1}:
\subsection{Clustering Predicted Probabilities}

For each class in the output layer of a neural network, we create \(M\) clusters of predicted probability values within the range [0, 1]. Each cluster represents a specific range of predicted probabilities and has a representative value, which can be the mean probability of the cluster. This clustering allows us to group predictions with similar confidence levels together, facilitating a more granular analysis of trustworthiness.

\subsection{Computing Trustworthiness Opinions}

For a given class \(c\) and for each cluster \(i\), we calculate the number of classifications \(n_i\) that fall into this cluster. We then compute the following:
\begin{itemize}
    \item \textbf{Positive Evidence:} The representative probability value of the cluster.
    \item \textbf{Negative Evidence:} This quantity is based on the difference between:
    \begin{enumerate}
        \item the ratio of the number of true classifications in the cluster ($t_c$) to the total number of predictions in the cluster ($n_c$) and
        \item the representative probability ($RP_i$).
    \end{enumerate}  The full equation is: 
    \begin{align}
        \alpha \mathbbm{1}_{t_c>n_c\times RP_i} (t_c - n_c\times RP_i) + \notag \\
    \beta \mathbbm{1}_{t_c<n_c\times RP_i}  (n_c\times RP_i - t_c) \notag
    \end{align}
    
     \(\alpha\) and \(\beta\) are chosen based on how much we want to penalize over and under-confidence.
    For simplicity, we use $\alpha=\beta=1$, thus the negative evidence is:
    
    $$\lvert t_c - n_c\times RP_i\rvert = n_c\lvert\frac{t_c}{n_c} - RP_i\rvert$$
    
    \item The total number of evidence in our framework is then: \[t_c + \lvert t_c - n_c\times RP_i\rvert = t_c(1+ \lvert 1 - \frac{n_c}{t_c}\times RP_i\rvert ) \] 
    which mostly depends on how far the the predictive probability is from actual probability. 
\end{itemize}
Using these evidences, we compute a subjective logic opinion for the trustworthiness of each probability cluster. This opinion incorporates degrees of belief, disbelief, and uncertainty, providing a nuanced measure of trustworthiness.

\subsection{Fusing Trust Opinions}

The trustworthiness opinions for each probability cluster are combined using appropriate fusion operators. This step allows us to synthesize the trust opinions into a single, comprehensive trust opinion that reflects the overall trustworthiness of the neural network’s probability predictions for class \(c\). The fusion process accounts for the varying levels of trustworthiness across different clusters, ensuring that the final trust opinion is balanced and representative.

After fusing trust opinion for each clusters, we obtain a trust opinion for the class. Then we have to fuse all these trust opinion (for each class) using another appropriate fusion operator to get the final trust opinion on the Neural Network.

The proposed trustworthiness quantification framework is designed to be computationally efficient. Its main steps—clustering predicted probabilities, computing trustworthiness opinions, and fusing them using subjective logic—operate with a complexity of \(\mathcal{O}(C \cdot M\cdot N)\) , where \(N\) is the number of predictions (i.e. number of data used for the quantification ), \(C\) is the number of classes, and \(M\) is the number of probability clusters. Since the complexity is linear, the framework scales efficiently and can be integrated into real-world applications without significant computational overhead.

Note that our work shares conceptual similarity with a paper on hallucination detection in LLMs using clustering of output entropy~\cite{Farquhar2024}. While that study targets semantic consistency, our method extends ECE with subjective logic to quantify trustworthiness.
\subsection{From Static to Dynamic}
To extend our framework from a static evaluation to a dynamic quantification of trustworthiness, we propose maintaining individual opinions for each probability cluster. During the operational phase, after inference, the predicted probabilities can be recorded and mapped to their corresponding clusters. Trustworthiness is then derived for each probability prediction within its respective cluster and label. These trustworthiness scores are subsequently fused using an appropriate fusion operator, dynamically evaluating the  trustworthiness of the output probabilities in real-time as new predictions are made.

\section{Experiments}
We evaluate two neural networks, one trained on the MNIST dataset and the other on CIFAR-10. The evaluation consists of assessing the trustworthiness of the models before and after calibration using temperature scaling. To quantify trustworthiness, we apply our algorithm (based on subjective logic) to calculate belief, disbelief, and uncertainty for individual probability clusters, and we use cumulative fusion to combine these opinions. We chose cumulative fusion because of its associativity and commutativity which allow us to generalize this fusion operators to more than two trustworthiness opinion~(Equation~\ref{eq:cumB}).
\begin{equation}
\left\{
\begin{aligned}
b_X^{(A \diamond B)} &= \frac{b_X^A u_X^B + b_X^B u_X^A}{u_X^A + u_X^B - u_X^A u_X^B}, \\
d_X^{(A \diamond B)} &= \frac{d_X^A u_X^B + d_X^B u_X^A}{u_X^A + u_X^B - u_X^A u_X^B}, \\
u_X^{(A \diamond B)} &= \frac{u_X^A u_X^B}{u_X^A + u_X^B - u_X^A u_X^B}, \\
a_X^{A \diamond B} &= \frac{a_X^A u_X^B + a_X^B u_X^A - (a_X^A + a_X^B) u_X^A u_X^B}{u_X^A + u_X^B - 2u_X^A u_X^B},
\\
&\text{if } u_X^A \neq 1 \lor u_X^B \neq 1, \\
a_X^{A \diamond B} &= \frac{a_X^A + a_X^B}{2}, 
\\
&\text{if } u_X^A = u_X^B = 1, \\
\end{aligned}
\right.
\label{eq:cumB}
\end{equation}
In our implementation, for simplicity, we always keep the number of positive and negative evidence allowing us to just add them at the end and then use Eq.\ref{eq:1} to derive the opinion since cumulative fusion is just an addition of evidence.

In this section, we describe the experimental setup used to train the neural network (NN) models. The setup is divided into four main components: datasets, model architectures, training procedure, and calibration method.
\subsection{Datasets}
We evaluate the models on two well-known datasets:
\paragraph{MNIST} The MNIST dataset consists of 70,000 grayscale images of handwritten digits (0–9), each of size 28×28 pixels. We split the dataset into 60,000 images for training and 10,000 for testing. The task is to classify each image into one of the 10 digit classes.
\paragraph{CIFAR-10} The CIFAR-10 dataset contains 60,000 color images of size 32×32, divided into 10 classes, with 6,000 images per class. The dataset is split into 50,000 training images and 10,000 test images. The classes represent various objects, such as airplanes, automobiles, birds, cats, and more. Each image is classified into one of the 10 object categories.

\subsection{Neural Network Architectures}
We designed two distinct neural network architectures, optimized for each dataset.

For the MNIST model, the input is a 28×28 grayscale image, flattened into a 1D array of 784 elements for the fully connected layers. A fully connected layer with 128 units and ReLU activation extracts 128 features, enabling the network to model complex relationships. The final fully connected layer with 10 units outputs logits for each digit class, which are converted to probabilities using the softmax function. This straightforward architecture effectively classifies MNIST digits due to the dataset's simplicity.


For CIFAR-10 we use a convolutional neural network (CNN) was used for CIFAR-10 classification. The architecture includes three convolutional layers with ReLU activation: the first uses 32 filters, and the next two use 64 filters, all with 3×3 kernels, to progressively extract features. After the first two convolutional layers, 2×2 max pooling reduces spatial dimensions and improves efficiency. The feature maps from the final convolutional layer are flattened into a 1D vector for the fully connected layers. A dense layer with 64 units and ReLU activation captures complex patterns, and a final softmax layer with 10 units outputs the probability distribution for CIFAR-10 classes. This design effectively handles the dataset's complexity, including variations in color, and shape.


Both neural networks were trained using the Adam optimizer, improving convergence speed. The MNIST model was compiled with the Adam optimizer, a sparse categorical crossentropy loss function. The model was trained for 100 epochs with a batch size of 32, using 90\% of the training data for training and 10\% as a validation set.

\subsection{Calibration: Temperature Scaling}

After training, we calibrate both models using temperature scaling, a post-processing technique designed to align predicted probabilities with actual outcomes. Temperature scaling modifies the model's output logits without changing its classification accuracy, thereby improving the reliability of the predicted probabilities.

In this method, a temperature parameter \(T\) is introduced to scale the logits output from the trained neural network. This adjustment is applied before the softmax function, which converts the logits into probabilities. The temperature parameter is optimized to minimize the negative log likelihood (NLL) loss, ensuring that the predicted probabilities better reflect the true likelihood of each outcome.

The calibration process involves an iterative optimization loop where the temperature parameter is fine-tuned over several calibration epochs. During each epoch, the model's logits are recalibrated by applying the temperature parameter, and the NLL loss is computed between the calibrated probabilities and the true labels. The gradients of the loss are then used to update the temperature parameter, reducing the difference between predicted probabilities and actual outcomes.

In the experiments, we used \(M=10\) groups for computing the trustworthiness.The clusters are \([\frac{i}{10}, \frac{i+1}{10}[\) for \(i =0, \ldots, 8\) with the final cluster covering \([\frac{9}{10}, 1]\)
The representatives are \(\frac{i}{10} + \frac{1}{20}\) for \(i =0, \ldots, 9\).

\section{Results and Discussion}

\begin{figure}
    \centering
    \includegraphics[width=\linewidth]{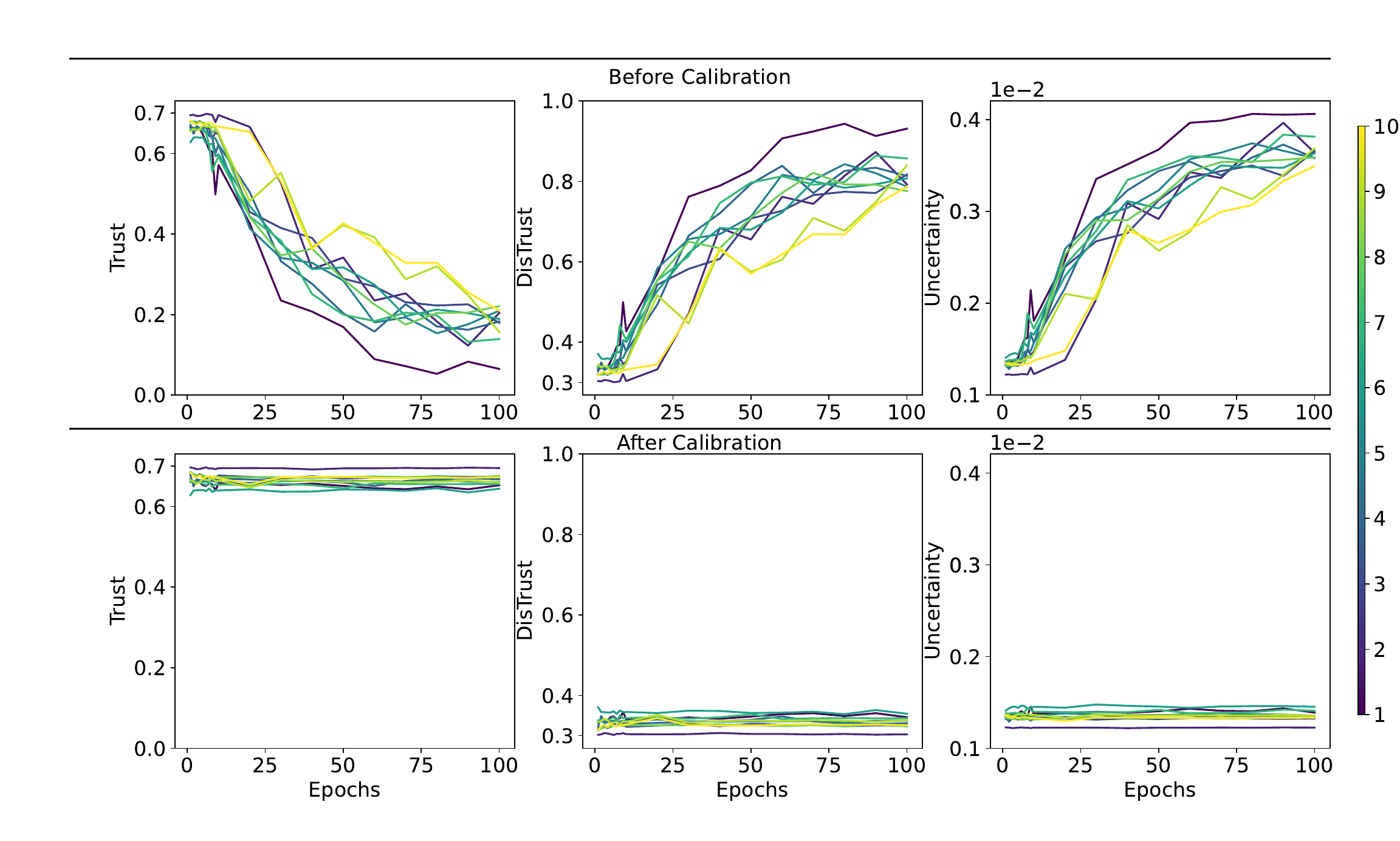}
    \caption{MNIST Result}
    \label{fig:mnist}
\end{figure}

\begin{figure}
    \centering
    \includegraphics[width=\linewidth]{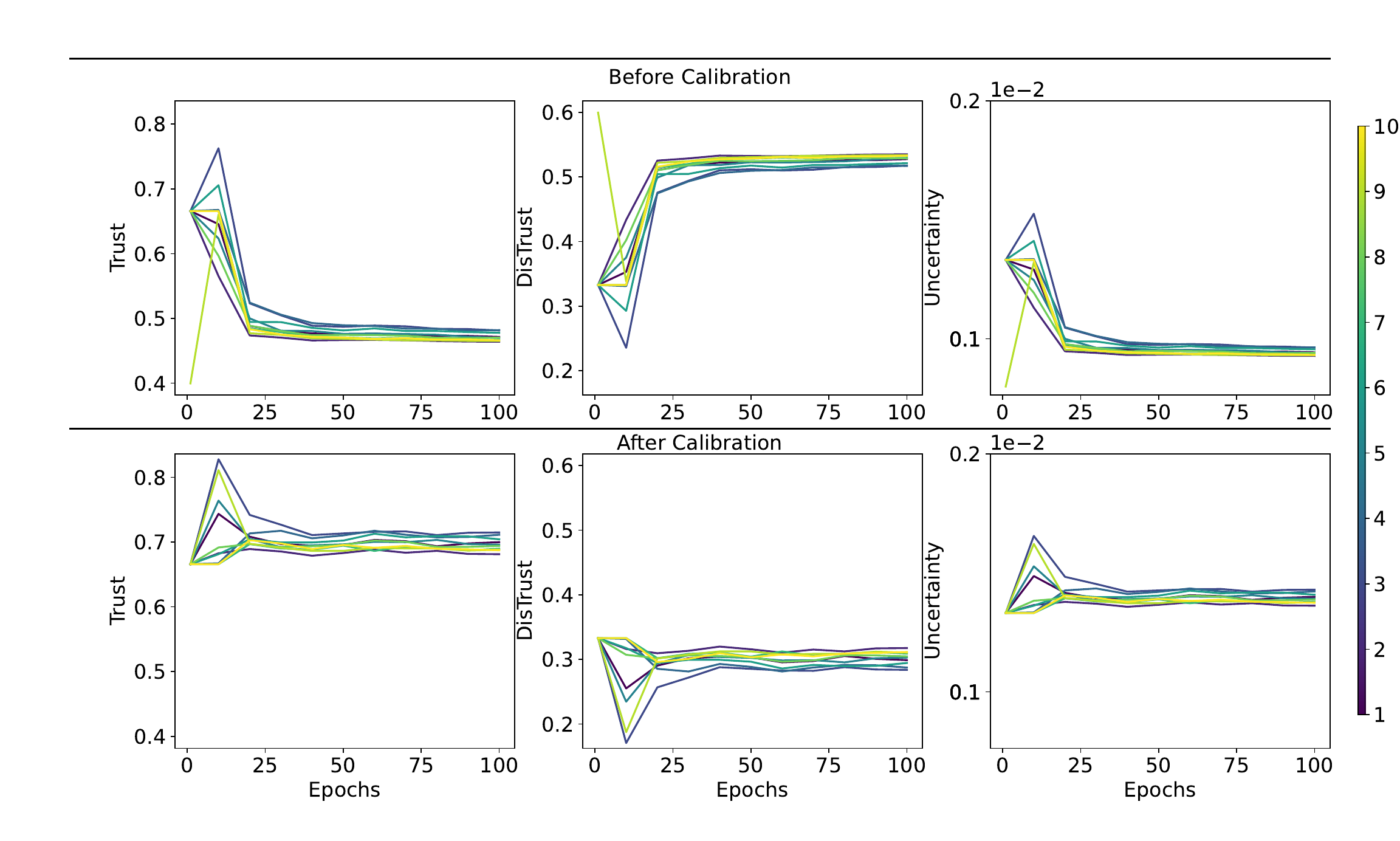}
    \caption{CIFAR 10 Result}
    \label{fig:cifar}
\end{figure}

This section compares the trustworthiness evaluation results before and after calibration (using temperature scaling) of the MNIST and CIFAR-10 neural networks. The metrics—trust, distrust, and uncertainty—were assessed over 100 epochs for all labels (0 to 9). The color bar in the figures represents the label corresponding to each curve, allowing us to observe the impact of calibration on individual labels. The results reveal distinct differences between the pre- and post-calibration performance, particularly in how the metrics evolve over time.

While the trustworthiness curves (trust/belief, distrust/disbelief, and uncertainty) start at similar points both before and after calibration, the post-calibration curves exhibit greater stability. In contrast, the pre-calibration performance tends to worsen, with a decrease in trust alongside an increase in distrust and uncertainty. 

\subsection{MNIST Results}
The results for the MNIST model are depicted in Figure~\ref{fig:mnist}.

\paragraph{Trust} Before calibration, trust started at a similar level as post-calibration (around 0.65), but as the training progressed, trust values decreased, fluctuating across different labels. This pattern suggests that the confidence in the network’s probabilities predictions diminished over epochs. In contrast, after calibration, trust remained constant. This demonstrates that temperature scaling helped the network maintain a higher and more consistent level of trust in its probabilities predictions.

\paragraph{distrust} Before calibration, distrust started low but increased as the number of training epochs grew, indicating growing overconfidence and/or underconfidence in predictions. However after calibration, distrust remained constant and stable at around 0.35. This improvement reflects the calibration’s effectiveness in reducing the network’s overconfidence and/or underconfidence in wrong predictions.

\paragraph{Uncertainty} Similarly, uncertainty began at a comparable point before and after calibration. However, without calibration, uncertainty tended to increase, showing less evidence during training. Post-calibration, uncertainty remained low and stable, indicating the model’s uncertainty stability.

\subsection{CIFAR10 Results}
For the CIFAR-10 dataset, a similar trend was observed, where the post-calibration model exhibited stable trustworthiness, while the pre-calibration model’s performance deteriorated over time. The results are illustrated in Figure~\ref{fig:cifar}

\paragraph{Trust} Both pre- and post-calibration trust curves began at similar values (at around 0.7) then increase a little bit for some labels to reach around 0.75 for pre calibration and 0.82 for post calibration. after that without calibration, as training continued, trust decreased to reach 0.48 after 20 epochs and remains constant. After calibration, trust decreases a little to come back to the initial value 0.7 after around 20 epochs and then remained constant. This reflects the impact of calibration in maintaining high confidence in probability predictions.

\paragraph{distrust} Both pre- and post-calibration distrust curves began at similar values, around 0.4. Before calibration, distrust gradually increased as training progressed, reaching a peak of 0.58 after approximately 20 epochs, where it remained relatively constant for the rest of the training. After calibration, distrust initially increased slightly, but then decreased steadily to return to the initial value of 0.4 around the 20th epoch. It remained stable thereafter. This shows that calibration was effective in controlling the network’s overconfidence in incorrect predictions, leading to more reliable performance over time.

\paragraph{Uncertainty} Pre- and post-calibration uncertainty curves also started at similar values, around 0.12. Before calibration, uncertainty increased slightly during the early training epochs, peaking at approximately 0.17 after 20 epochs, and remained at that level for the rest of the training process. After calibration, uncertainty initially increased similarly, but quickly dropped back to the original level of 0.12 around epoch 20 and remained constant thereafter. This demonstrates that calibration helped the model to reduce uncertainty, particularly in later epochs, resulting in more stable and confident predictions.

Overall, the results show that temperature scaling significantly improved the stability of the trustworthiness metrics, maintaining higher trust and lower distrust and uncertainty, compared to the pre-calibration network, where trust diminished and distrust and uncertainty increased as training progressed.

\subsection{Discussion}



\begin{figure}
    \centering
    \begin{subfigure}{0.23\textwidth}
    \centering
    \includegraphics[width=\textwidth]{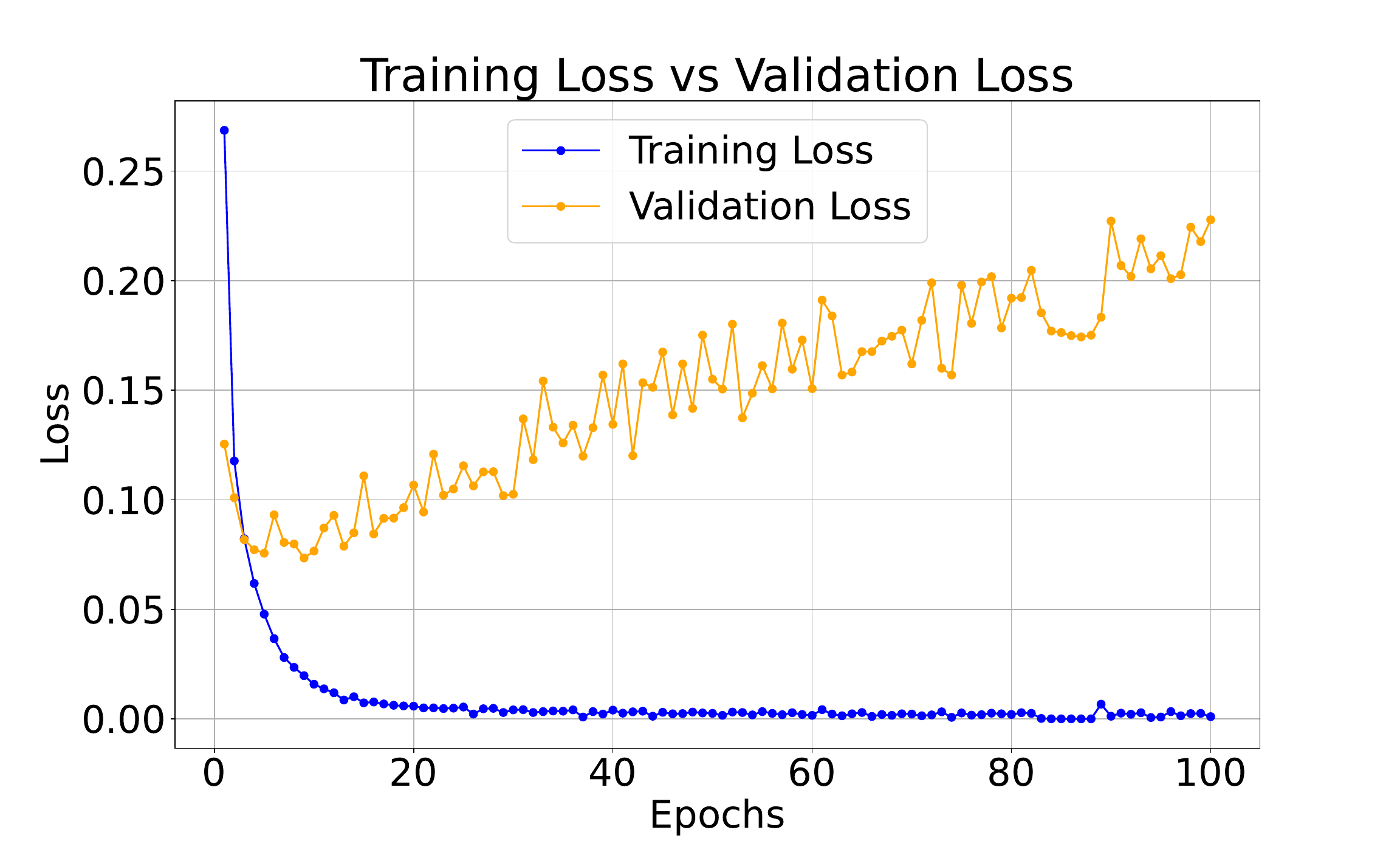}
    \caption{Training and Validation Loss of MNIST model During training}
    \label{fig:mnistovfit}
    \end{subfigure}
    \hfill
    \begin{subfigure}{0.23\textwidth}
    \centering
    \includegraphics[width=\textwidth]{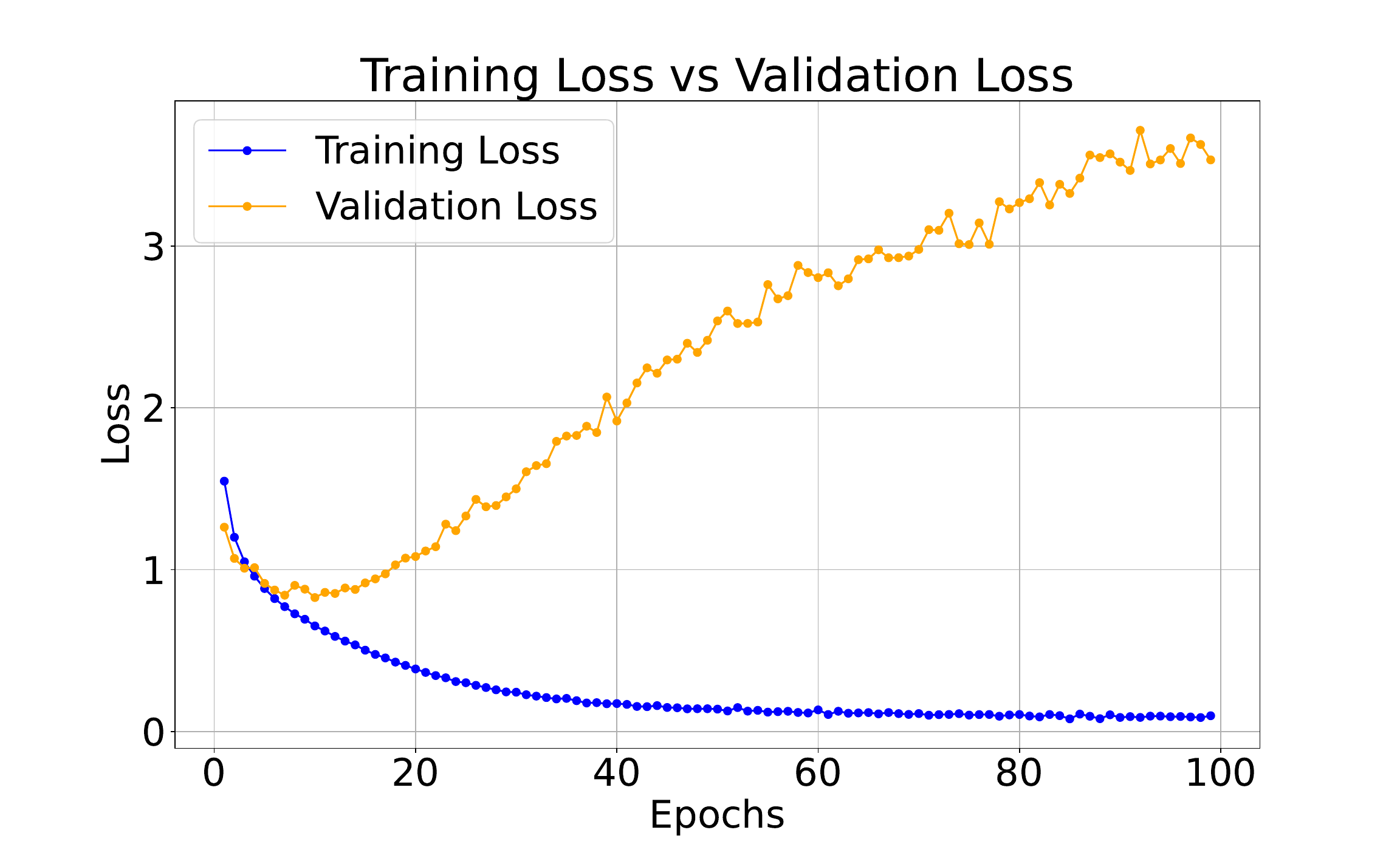}
    \caption{Training and Validation Loss of CIFAR10 model During training}
    \label{fig:cifarovfit}
    \end{subfigure}
    \label{fig:ovfit}
\end{figure}

The behavior of the metrics for the CIFAR-10 model can be explained by the fact that the model requires around 10 epochs to be well-trained. After this point, as shown in Figure~\ref{fig:cifarovfit}, the model begins to overfit, as indicated by the increasing gap between the training and validation losses. This overfitting leads to poor generalization and explains why the metrics—trust, distrust, and uncertainty—are optimal for both pre- and post-calibration models during the early epochs, but deteriorate sharply for the pre-calibration model after about 10 epochs. Specifically, trust decreases, and distrust and uncertainty increase for the pre-calibration model as overfitting sets in. In contrast, the post-calibration model experiences a slight decline in performance but stabilizes quickly, maintaining higher trustworthiness throughout the training.
\paragraph{Overfitting in CIFAR-10} The overfitting observed in the CIFAR-10 model results in a more overconfident network. This overconfidence leads to poor probability predictions, which causes trust to decrease and distrust to increase, as evidenced by the worsening of these metrics in the pre-calibration model. After around 20 epochs, trust in the pre-calibration model drops to around 0.48, and distrust peaks at 0.58, remaining relatively constant thereafter (as shown in the results).
\paragraph{Post-Calibration Stability} Temperature scaling mitigates the effects of overfitting by adjusting the model’s confidence levels. This explains why the post-calibration model remains stable after the initial performance dip and avoids the severe deterioration seen in the pre-calibration model. After calibration, trust decreases only slightly and stabilizes around 0.7 after 20 epochs, while distrust and uncertainty also stabilize at much lower levels compared to the pre-calibration model.

For the MNIST model, the dynamics exhibit similarities to those observed in CIFAR-10 but with a less pronounced effect. As shown in Figure~\ref{fig:mnistovfit}, the gap between training and validation losses increases throughout training, though not as significantly as in CIFAR-10. Toward the later epochs, the gap appears to stabilize, suggesting a lower degree of overfitting compared to CIFAR-10. This explains why we do not observe the same sharp pattern of improvement followed by deterioration seen in CIFAR-10, where overfitting is more severe.

\paragraph{MNIST Training Behavior } The MNIST model reaches its best performance relatively early in training, and there is little difference in behavior between the pre- and post-calibration models. The absence of significant overfitting means that trust remains relatively constant throughout training for both models, and distrust and uncertainty are minimal. The difference between the pre- and post-calibration models is less pronounced because the MNIST task is simpler and requires fewer epochs to reach optimal performance.

\paragraph{Impact of Overfitting on Trustworthiness }Overfitting leads to overconfident neural networks, which in turn produce poorly calibrated probability predictions. This decreases belief (or trust) in the network’s predictions and increases distrust and uncertainty. In the CIFAR-10 model, these effects are evident, with trust decreasing and distrust increasing as overfitting sets in. However, temperature scaling corrects for this by recalibrating the model’s confidence, which helps the post-calibration model remain trustworthy even after the base model begins to overfit.


\section{Conclusion}
In this paper, we presented a novel framework for quantifying the trustworthiness of neural networks, particularly in classification tasks, by leveraging subjective logic. Our approach extends traditional Expected Calibration Error (ECE) by incorporating belief, disbelief, and uncertainty, providing a more interpretable and nuanced evaluation of model trustworthiness. Through clustering predicted probabilities and fusing trustworthiness opinions, our method offers a comprehensive, intuitive way to assess the reliability of neural networks, addressing key limitations in existing calibration techniques.

The evaluation of our framework demonstrated its effectiveness in maintaining stable trustworthiness over time, particularly when applied to calibrated models. Our experimental results on MNIST and CIFAR-10 datasets highlight the improvements in trust, distrust, and uncertainty metrics after calibration, showcasing the utility of temperature scaling in enhancing model reliability. By dynamically updating trustworthiness during inference, this framework has the potential to contribute to the ethical deployment of AI systems in critical domains such as healthcare and autonomous systems.


\emph{Future Works.} 
Future work will focus on elaborating the scheme to ensure that uncertainty is solely dependent on the amount of data. This would guarantee that uncertainty reflects the quantity of available evidence, preventing redundancy with disbelief. In line with recent advances on dataset trustworthiness quantification, which introduced three distinct opinion quantification schemes for handling uncertainty under different evidence conditions \cite{ouattara2025assessingtrustworthinessaitraining}, we will investigate how similar strategies can be integrated into our framework. Another direction is to enhance the dynamic aspects of the framework to enable real-time trustworthiness updates during the operational phase. Finally, we aim to study the impact of the number of clusters on the results, as more clusters allow for greater dynamism but reduce the amount of evidence available for each cluster.


\bibliographystyle{plain} 
\bibliography{conference_101719}

\end{document}